\documentclass[conference]{IEEEtran}
\usepackage{graphicx,times,amsfonts,amssymb,amsmath,multirow,subfigure} % Add all your packages here
\usepackage{algorithmic,algorithm,url}
\usepackage{tikz}
\usepackage{textcomp}
\usepackage{hyperref}
\usepackage{lipsum}

\newcommand\copyrighttext{%
  \footnotesize \textcopyright 2018 IEEE. Personal use of this material is permitted.
  Permission from IEEE must be obtained for all other uses, in any current or future
  media, including reprinting/republishing this material for advertising or promotional
  purposes, creating new collective works, for resale or redistribution to servers or
  lists, or reuse of any copyrighted component of this work in other works.\\
  \emph{To appear in the Proceedings of the 2018 International Joint Conference on Neural Networks (IJCNN 2018)}}
\newcommand\copyrightnotice{%
\begin{tikzpicture}[remember picture,overlay]
\node[anchor=north,yshift=-10pt] at (current page.north) {\fbox{\parbox{\dimexpr\textwidth-\fboxsep-\fboxrule\relax}{\copyrighttext}}};
\end{tikzpicture}%
}

\newcommand{\W}{\mathbf{W}}

\newcommand{\Wout}{\W_{out}}

\IEEEoverridecommandlockouts    % to create the author's affliation portion
\begin{document}

% paper title: Must keep \ \\ \LARGE\bf in it to leave enough margin.

\title{Concentric ESN: Assessing the Effect of Modularity in Cycle Reservoirs}
\author{\IEEEauthorblockN{Davide Bacciu and Andrea Bongiorno}
\IEEEauthorblockA{Dipartimento di Informatica\\
Universit\`a di Pisa\\
Largo B. Pontecorvo 3\\
Pisa, Italy\\
Email: bacciu@di.unipi.it}
}

% avoiding spaces at the end of the author lines is not a problem with
% conference papers because we don't use \thanks or \IEEEmembership
%\specialpapernotice{(Invited Paper)}

\maketitle
\copyrightnotice

\begin{abstract}
The paper introduces concentric Echo State Network, an approach to design reservoir topologies that tries to bridge the gap between deterministically constructed simple cycle models and deep reservoir computing approaches. We show how to modularize the reservoir into simple unidirectional and concentric cycles with pairwise bidirectional jump connections between adjacent loops. We provide a preliminary experimental assessment showing how concentric reservoirs yield to superior predictive accuracy and memory capacity with respect to single cycle reservoirs and deep reservoir models.
\end{abstract}

% no key words
\IEEEpeerreviewmaketitle

%%% PER CAMERA READY:  TREE ESN COME COSA INTERESSANTE E DROPIN + NEUROCOMPBIOBEATS IN APPLICAZIONI INTRODUZIONE.

\section{Introduction} \label{sect:intro}
Reservoir Computing (RC) \cite{Lukosevicius2009} is a paradigm for recurrent neural network design that, in the latter years, has found wide application thanks to its (relative) simplicity and effectiveness in dealing with sequential data processing tasks, such as with sensor network data \cite{Bacciu2013,dropin},mobile robot navigational data \cite{antonelo,syroco}, telephone load forecasting \cite{tel}, ambient assisted living \cite{Dragone2015269,Amato2015}, biomedical data\cite{berg,Bacciu20189}, etc. RC is based on a conceptual separation between the recurrent part of the network, i.e. the \emph{reservoir}, and the feedforward part, including the input and the output layers. The reservoir layer acts as a dynamic memory of the history of the input signals. The activations of its neurons are then read and combined by the output layer, which is known also as readout, to compute the network prediction.

Echo State Networks (ESNs) \cite{Jaeger2004} are by far the most popular RC model and are characterized by the fact that the readout layer is, typically, the only trained part of the network. The input and reservoir connections, instead, are randomly initialized, under conditions ensuring contractivity \cite{JaegerTech2001}, and then left untrained. This ESN feature, on the one hand, enables efficient training as the readout neurons are typically linear units whose weights can be determined by batch least square minimization. On the other hand, it poses much attention on how the input and reservoir weights are initialized, as this induces a strong architectural bias that is not compensated by training.

The latter aspect has generated a large body of works studying various architectural and topological features and the respective capacities in terms of memory size, speed of forgetting and ability to model short and long term dependencies in the input signal \cite{jaeger2002}. Initially, much of the attention focused on randomly generated reservoirs characterized by sparse connectivity, where the degree of sparsity, the size of the reservoir, as well as the spectral properties of the reservoir weight matrix are model hyperparameters. As such, much of the computational cost of obtaining a "good" ESN predictor shifted from the actual training of the readout, to the potentially combinatorial cost of the model selection procedure, although some works \cite{practical} have provided practical hints to restrict the choice of the hyperparameters' intervals.

A different approach to control model selection cost has been put forward by \cite{scr} which proposes to limit the effect of randomization in the ESN construction by using simple deterministically built reservoirs, with strongly tied weights both for the input and the reservoir connections. Among the different reservoir topologies explored in \cite{scr}, the simple cycle reservoir, i.e. where neurons are organized into an unidirectional loop where all connections have the same weight, has proved the best tradeoff between simplicity and memory capacity. This simple cycle reservoir has been later extended with shortcut connections which basically allow to shorten the path between two neurons in the loop by means of bidirectional cross-loop connections \cite{crj}. This latter model has shown to be competitive with respect to standard ESNs with randomly and sparsely connected reservoirs. With respect to such constrained deterministic architectures, the Deep ESN model proposed in \cite{deepESN} positions on the opposite side of the spectra in terms of architectural complexity. There it is shown how certain modularization and layering strategies typical of deep learning can be effectively applied to reservoir design, yielding to excellent results in terms of predictive and memory capacity.

In this paper, we put forward a novel reservoir design that tries to borrow the best of the two worlds, i.e. the deterministic cycle reservoir and the deep-style modularization. We introduce the idea of a concentric reservoirs, where the neurons are organized into a number of simple unidirectional cycles of various length, represented concentrically for graphical clarity. Each of such cycle might be characterized by a different connection weight and it has full connectivity from the input units and towards the readout neurons. In addition, we will consider the effect of having bidirectional jump connections between the loops.

The rationale for the concentric ESN model is to enhance the reservoir dynamic memory with the ability to explicitly encode paths of different length, temporal information at different timescales and with different fading memories of the past inputs, by acting on the loop lengths and on weight initialization. Jump connections, in this sense, are expected to provide the ability to switch between different memory regimes represented by the different loops. The aim of this paper is that of providing a preliminary characterization of effectiveness of the concentric architecture, studying the effect of various topological factors (such as jumps and cycles size) in terms of predictive accuracy as well as of empirical memory capacity \cite{jaeger2002}.

The remainder of the paper is organized as follows: Section \ref{sect:back} summarizes the ESN model and the baseline deterministic single cycle reservoir approaches. Section \ref{sect:cesn} introduces the concentric ESN model, while Section \ref{sect:expcomp} provides a comparative experimental analysis of its performance with respect to single cycle models and the deepESN \cite{deepESN}. Finally, Section \ref{sect:conclude} concludes the paper.

\section{Background} \label{sect:back}
In this section, we provide a brief overview of the Echo State Network model, in Section \ref{sect:esn}, and we discuss in Section \ref{sect:cr} two of its variants that introduce deterministically initialized reservoir topologies.

\subsection{Echo State Networks} \label{sect:esn}
\begin{figure}[tb]
  \centering
  \includegraphics[width=.8\columnwidth]{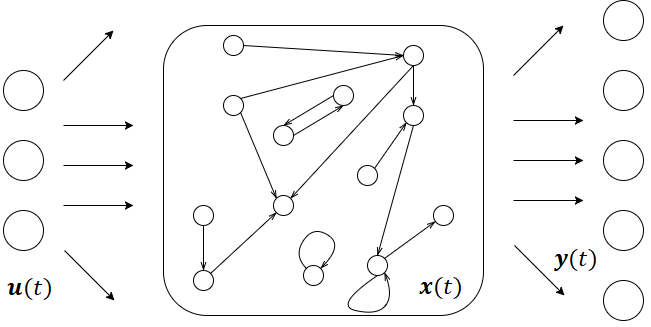}
  \caption{Architecture of a standard ESN comprising an input layer, a sparsely connected recurrent reservoir layer and the readout layer with full connectivity from the reservoir.}\label{fig:ESN}
\end{figure}

Echo State Networks (ESNs) \cite{Jaeger2004} are the most popular model in the reservoir computing paradigm, due to their conceptual and implementation simplicity, coupled with effectiveness on many sequential data processing tasks and thanks to the efficiency in training. An ESN, such as the one illustrated in Figure \ref{fig:ESN},
comprises an input layer with $N_U$ units, a reservoir layer with $N_R$ units and a readout with $N_Y$ units.

The reservoir is a large, sparsely-connected layer of recurrent non-linear units (typically $tanh$) which is used to perform a contractive encoding \cite{JaegerTech2001} of the history of driving input signals into a state space. The readout comprises feedforward linear neurons computing ESN predictions as a weighted linear combination of the reservoir activations. A variant of the standard ESN that is very popular is the leaky integrator ESN (LI-ESN) which applies an exponential moving average to the reservoir state space values. This allows a better handling of input sequences that change slowly with respect to sampling frequency \cite{Lukosevicius2009} and it has been shown to work best, in practice, when dealing with sensor data streams \cite{bacciu2013experimental}.
At each time step $t$, the reservoir activation of a LI-ESN is computed by
\begin{equation}
\label{eq.reservoir}
\mathbf{x}(t) = (1 - \alpha) \mathbf{x}(t - 1) + \alpha f(\mathbf{W}_{in} \mathbf{u}(t) + \mathbf{W}_h \mathbf{x}(t - 1))
\end{equation}
where $\mathbf{u}(t)$ is the vector of $N_U$ inputs at time $t$, $\mathbf{W}_{in}$ is the $N_R \times N_U$ input-to-reservoir weight matrix, $\mathbf{W}_h$ is the $N_R \times N_R$ recurrent reservoir weight matrix and $f$ is the component-wise reservoir activation function.  The term $\alpha \in [0,1]$ is a \emph{leaking rate} which controls the speed of LI-ESN state dynamics, with larger values denoting faster dynamics.

Reservoir parameters are left untrained after a random initialization subject to the constraints given by the so called \emph{Echo State Property} (ESP) \cite{JaegerTech2001}, requiring that network state asymptotically depends on the driving input signal and any dependency on initial conditions is progressively lost. In  \cite{JaegerTech2001}, it is provided a necessary and a sufficient condition for the ESP. The sufficient condition states that the largest singular value of the reservoir weight matrix (or its leaky integrated variant in LI-ESN) must be less than $1$. The necessary condition \cite{JaegerTech2001}, on the other hand, says that if the \emph{spectral radius} $\rho(\tilde{\mathbf{W}})$, i.e. the largest absolute eigenvalue of the matrix $\tilde{\mathbf{W}}$, is larger than $1$, the network as an asymptotically unstable null states and hence lacks the ESP. The sufficient condition is considered by to be too restrictive \cite{JaegerTech2001} for practical purposes, whereas the reservoir matrix is often initialized to satisfy the necessary condition, i.e.
\begin{equation}
\label{eq.initialization}
\rho(\mathbf{W}_R) < 1,
\end{equation}
with values of the spectral radius that are, typically, close to the stability threshold $1$. Input weights are randomly chosen from a uniform distribution over $[-s_{in}, s_{in}]$ (where $s_{in}$ is an input scaling parameter), while $\mathbf{W}_h$ is typically from a uniform distribution in $[-1,1]$ and then scaled so that Eq.~\ref{eq.initialization} holds.

The ESN output is computed by the readout through the linear combination
\begin{equation}
\label{eq.readout}
\mathbf{y}(t) = \mathbf{W}_{out} \mathbf{x}(t)
\end{equation}
where $\mathbf{W}_{out}$ is the ${N_Y \times N_R}$ reservoir-to-readout weight matrix. Training of an ESN model amounts to learning the values of the $\Wout$ matrix which implies the solution of a linear least squares minimization problem. The approach typically used to solve this problem employs ridge regression \cite{Lukosevicius2009}, where the output weight matrix is obtained as
\begin{equation}
\label{eq.ridge}
\mathbf{W}_{out} = \mathbf{Y}_{true} \mathbf{X}^T \left(\mathbf{X}\mathbf{X}^T + \lambda \mathbf{I}\right)^{-1}
\end{equation}
where $\mathbf{Y}_{true}$ is the $N_Y \times T$ matrix of true outputs for the $T$ samples,  $\mathbf{X}$ is the $N_R \times T$ matrix of readout states and $\mathbf{I}$ is the identity matrix. The term $\lambda$ is the (model-selected) ridge regression parameter controlling the amount of regularization applied.

\subsection{Cycle Reservoir Architectures} \label{sect:cr}
The ESNs are fairly efficient when it comes to training, given the relative simplicity of solving the linear least squares minimization problem required to find the readout weights. On the other hand, they require a lengthy and careful model selection process, as the identification of the most suited reservoir configuration (connectivity and weights) for a given task typically requires to explore several randomly chosen configurations.

\begin{figure}[tb]
  \centering
  \subfigure[SCR\label{fig:scr}]{\includegraphics[width=.8\columnwidth]{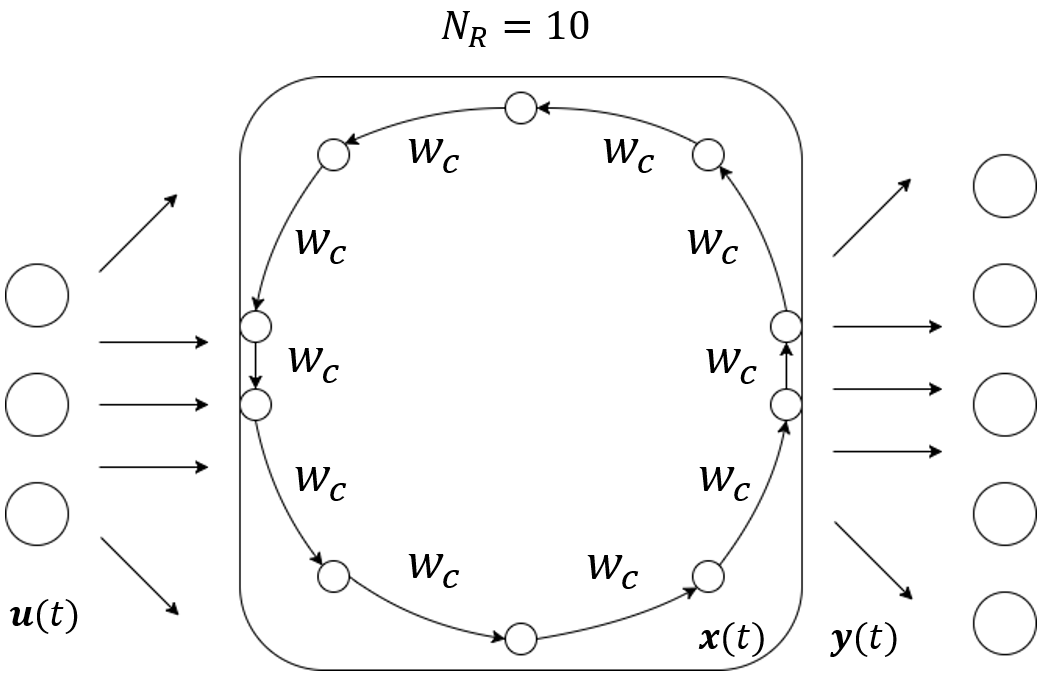}}
  \subfigure[CRJ\label{fig:crj}]{\includegraphics[width=.8\columnwidth]{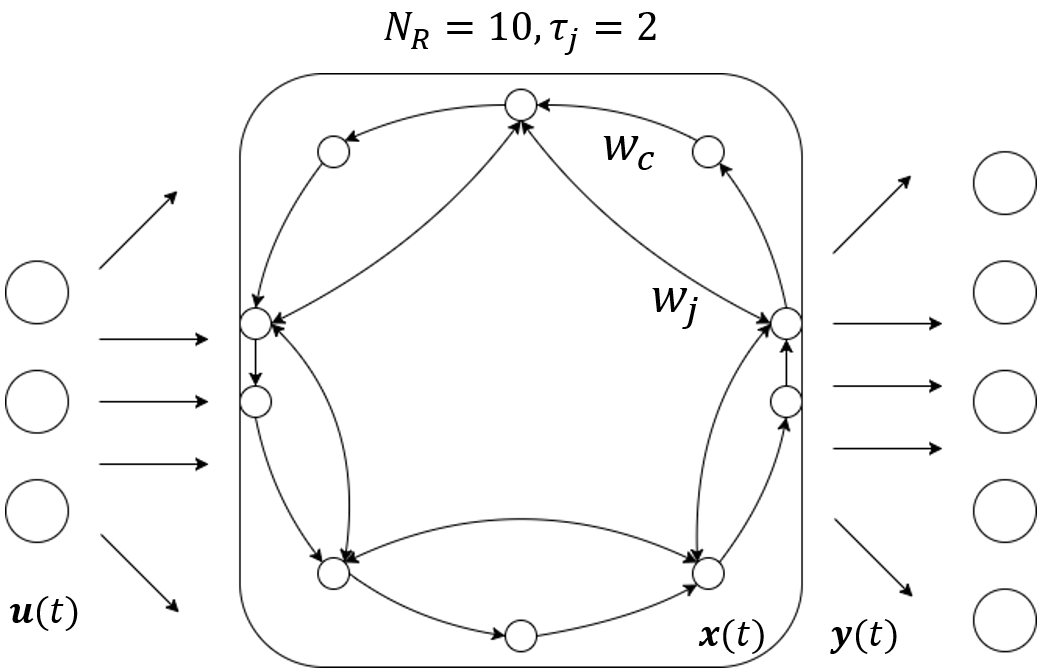}}
  \caption{Example of an (a) SCR with $N_R=10$ reservoir neurons and (b) the corresponding CRJ with jump step $\tau_j=2$.}\label{fig:crESN}
\end{figure}

In an attempt to control the effect of randomization in the construction of the reservoir, a number of papers have proposed deterministic approaches to construct the ESN reservoir. These are based on a simplification of the reservoir topology to a single loop structure which has been first introduced in \cite{scr} under the name of Simple Cycle Reservoir (SCR). Figure \ref{fig:scr} shows an exemplar SCR network: like a standard ESN it includes an input layer with untrained weights and a readout with adaptive weights, both fully connected to the reservoir neurons. The reservoir is arranged in a uni-directional loop where each neuron receives exactly one input from its predecessor in the cycle and all reservoir connections have the same weight $w_c$. Similarly, all input weights have the same absolute value $v$ with the sign being generated deterministically from a single aperiodic sequence, e.g. using a decimal expansion of the $\pi$ irrational number (see \cite{scr} for details).

The SCR model has been introduced mainly to study theoretical properties of simple and deterministically constructed reservoirs. In this sense, its predictive performance does not match that of a standard randomly constructed ESN \cite{scr}. In \cite{crj}, it has been proposed a generalization of the SCR capable of
reducing the performance gap with respect to standard ESN, eventually surpassing it on several timeseries prediction benchmarks. This model, referred to as Cycle Reservoir with Regular Jumps (CRJ), introduces bi-directional shortcut connections (jumps) in the structure of a uni-directed SCR loop as depicted in Figure \ref{fig:cjESN}. The rationale for the introduction of jump connectivity is that of reducing the clustering degree and the average path length of the reservoir, in accordance with the foundational ideas of reservoir computing \cite{crj}.

The shortcut connections have all the same weight $w_j$ which is chosen independently from the main cycle weight $w_c$: the remainder of the reservoir weights are clearly set to zero. The introduction of shortcut connection brings in an additional hyperparameter $\tau_j$ that defines the length of the jump, whose interpretation can be clearly understood in Figure \ref{fig:cjESN}. In practice, the jump step parameter determines that there exist a shortcut connection between two reservoir neurons every $\tau_j$ positions in the reservoir cycle, while the remainder of the neurons have only unidirectional loop connections.

In this paper, we put forward the idea that the simple cycle reservoirs can be further extended to accommodate for multiple cycles in a modular organization of the reservoir that allows capturing a larger variety of dynamical time scales, while maintaining advantages in terms of controlling randomization effects in reservoir initialization.

\section{Concentric Echo State Networks} \label{sect:cesn}

\begin{figure}[tb]
  \centering
  \includegraphics[width=.8\columnwidth]{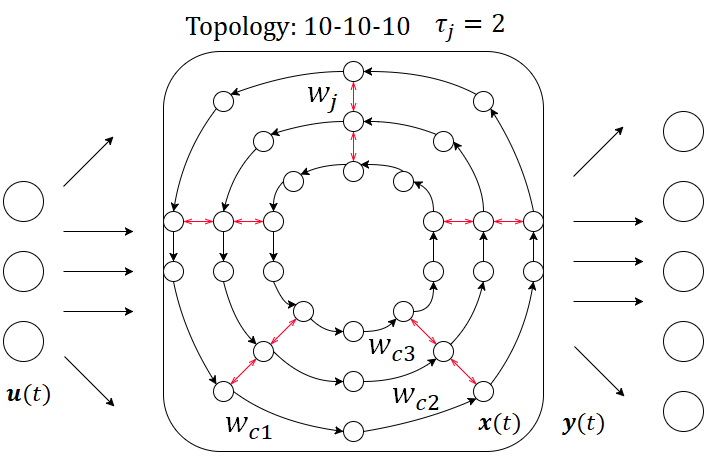}
  \caption{Architecture of a concentric network including jumps (cjESN), with $3$ reservoir modules, a total of $N_{R} = 30$ reservoir nodes, and jump size $\tau_j = 2$. Note that this network has, formally, a single reservoir of 3 circular modules connected by the bi-directional jump links (in red in the image). Each node in the reservoir is thus connected to the readout and receives input from the input neurons. A cESN has the same architecture without the jump connections.}\label{fig:cjESN}
\end{figure}

We propose {\it Concentric Echo State Networks} as a natural generalization of the deterministic reservoir models discussed in Section \ref{sect:cr}. The concentric ESN has a modular reservoir architecture comprising a number of uni-directional cycle reservoirs, each fully connected both to the input and the output neurons. Following the SCR-CRJ distinction, we consider two broad classes of concentric architectures: the first is characterized by independent reservoir cycles which are disconnected one-another and that is referred to as concentric ESN (cESN). The second architecture allows the presence of bi-directional jump connections between loops and, in the following, it is referred to as cESN with jumps (cjESN). Figure \ref{fig:cjESN}  provides a graphical illustration of the proposed cjESN model, with an exemplar reservoir comprising $3$ modules/cycles of $10$ reservoir each. Jump connections are highlighted in different color (red): the corresponding cESN architecture can straightforwardly be obtained from Figure \ref{fig:cjESN} by removing the jump connections. As in the deterministic architectures connections in a loop $i$ share the same weight $w_{ci}$, while jump connections have weight $w_j$ (which can in principle vary between couples of connected loops). As in CRJ we define a jump size topological parameter $\tau_j$ which controls how often we place the jump connections between loops (again, also this parameter can be made dependent on the couple of connected cycles).

The reservoir of the cjESN is organized into concentric loops for graphical clarity, but the general model does not imply a specific ordering between the modules induced by cycle length. Nevertheless, as the experimental analysis will show, we expect that the presence of such an ordered organization of the cycles would lead to better efficacy of the model. The intuition underlying the cjESN architecture is that of introducing the possibility of capturing a large variety of dynamical time scales by having cycle paths of different length, i.e. different number of neurons. In practice, by having cycles of different length we are doing explicitly what the CRJ model is doing by means of the jump connections, which allow to shorten the unique long cycle of the CRJ reservoir by taking shortcuts. On the other hand, the cjESN allows to explicitly include paths of different lengths as different cycles of the reservoir. Jump connections, in such a scenario, allow to move between different timescales, even between the same timeseries, by jumping on the next/previous level cycles. Even richer dynamics can be envisaged, such as having bottleneck cycles as intermediate modules of the reservoir, creating points of strong degree of local clustering, which can be a desired behaviour in certain tasks.

\begin{figure}[tb]
  \centering
  \includegraphics[width=.92\columnwidth]{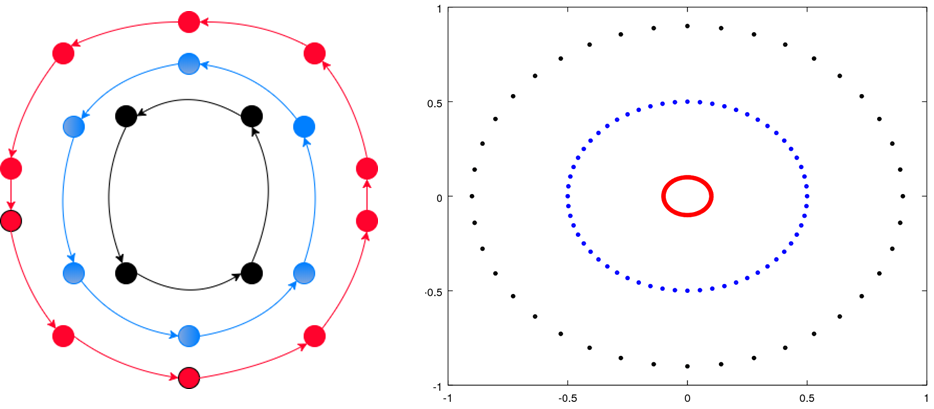}
  \caption{A graphical comparison of concentric reservoirs (left) and the eigenvalues of the corresponding reservoir matrices (right). Note that, to simplify plots, the concentric cycles on the left display only one tenth of the actual neurons used to plot the eigenvalues on the right.}\label{fig:eig}
\end{figure}

The use of a modular architecture has an additional advantage, which comes at the cost of a little increase in the complexity of the model selection procedure. It allows to have loops with different cycle weights $w_{ci}$, as depicted in Figure \ref{fig:cjESN}. As highlighted in \cite{crj}, a cycle reservoir is characterized by the spectral property of having the eigenvalues of the reservoir matrix organized into a cycle of radius $w_c$, i.e. the value of the cycle weight. This property extends to the concentric model where, as highlighted in Figure \ref{fig:eig}, the eigenvalues of the reservoir matrix organize in concentric cycles of radius $w_{ci}$, where the subscript $i$ denotes the $i$-th cycle. Such a property is quite desirable for reservoir design as, taken together with cycle length, it allows to gauge both the memory path length, through the number of neurons in a cycle, as well as memory speed, in terms of how fast the memory of an input fades in the cycle which is influenced by the spectral radius of the module.  Figure \ref{fig:eig} shows a paradoxical configuration where the largest reservoir module (the most external one, in red in figure) is the one with the smallest spectral radius (e.g. modeling transition between sever short-term subtasks), while the smallest innermost cycle (in black) has the largest spectral radius (modeling longer term dependencies over relatively short paths). Jump connections, in this sense, allow to switch between this different dynamical memories of different capacity and speed. They perturb the circle-like spectral organization of the reservoir by adding points within the unitary sphere but outside of the circles perimeter.

Training of a concentric ESN can be performed as with the standard ESN in Section \ref{sect:esn}. The concentric configuration of the reservoir has only effect on the structure of the reservoir weight matrix. For clarity, we can assume reservoir nodes to be indexed following the ordering in the unidirectional cycle they belong to;  cycles, instead, are ordered from the outermost to the innermost in the concentric structure. In this case, a cESN reservoir weight matrix has only zeros on the main diagonal while it is organized in blocks of lower diagonal stripes, corresponding to cycle loops, plus some off diagonal elements for closing the unidirectional cycles. The lower diagonal of the reservoir weight matrix has only zeros in those positions where one cycle ends and the next cycle begins. A cjESN has the same structure plus off-diagonal elements for the jump weights (on both upper and lower triangular, since jumps are bi-directional). For instance, if we consider a cjESN with two cycles of $4$ and $6$ neurons, respectively, and a jump size of $\tau_j = 3$, the reservoir weight matrix would look like the following one:
\[
\left(
  \setlength\arraycolsep{2pt}
  \begin{array}{cccccccccc}
    0 & 0 & 0 & w_{c1} & w_j & 0 & 0 & 0 & 0 & 0 \\
    w_{c1} & 0 & 0 & 0 & 0 & 0 & 0 & 0 & 0 & 0 \\
    0 & w_{c1} & 0 & 0 & 0 & 0 & 0 & 0 & 0 & 0 \\
    0 & 0 & w_{c1} & 0 & 0 & 0 & 0 & w_j & 0 & 0 \\
    w_j & 0 & 0 & 0 & 0 & 0 & 0 & 0 & 0 & w_{c2} \\
    0 & 0 & 0 & 0 & w_{c2} & 0 & 0 & 0 & 0 & 0 \\
    0 & 0 & 0 & 0 & 0 & w_{c2} & 0 & 0 & 0 & 0 \\
    0 & 0 & 0 & w_j & 0 & 0 & w_{c2} & 0 & 0 & 0 \\
    0 & 0 & 0 & 0 & 0 & 0 & 0 & w_{c2} & 0 & 0 \\
    0 & 0 & 0 & 0 & 0 & 0 & 0 & 0 & w_{c2} & 0 \\
  \end{array}
\right).
\]

In the remainder of this paper, we will consider concentric ESN models with the same jump size and jump weight for all the concentric cycles. For the sake of simplicity of the model selection phase, in this preliminary assessment of the concentric ESN architecture we consider all reservoir cycles to have the same weight $w_c$. Training of the model is performed using ridge regression as in Eq. (\ref{eq.ridge}).

\section{Experimental Results} \label{sect:expcomp}
We provide an experimental assessment of the effect of the concentric reservoir topology with and without jumps, comparing its performance with that of the two reference cycle reservoir networks in literature, i.e. SCR \cite{scr} and CRJ \cite{crj}. To this end we consider two timeseries prediction tasks from the original SCR and CRJ papers, whose results are discussed in Section \ref{sect:time}. These papers also report performance of a standard ESN, for comparison. This analysis is complemented, in Section \ref{sect:mc}, with an empirical study of the memory capacity of the newly introduced concentric reservoirs.

\subsection{Timeseries Prediction Tasks} \label{sect:time}
The first task is a standard system identification problem, referred to as NARMA, consisting in the identification of the 10-th order nonlinear autoregressive moving average model \cite{narma} given by
\[
y(t+1) = 0.3 y(t) + 0.05 y(t) \sum_{i=0}^{9} y(t-i) + 1.5 s(t-9) s(t) + 0.1
\]
where $y(t)$ is system output at time $t$ and $s(t)$ is the corresponding state generated uniformly in $[0,0.5]$. The predictive task requires to estimate $y(t)$ using as input the current state $s(t)$. The dataset consists of a NARMA sequence, normalized in $[0,1]$ with a length of $10000$ samples, such that the first $2000$ items have been used as training set, the following $5000$ as validation set and the remainign $3000$ as test set.

The second task is based on the Santa Fe Laser dataset \cite{Jaeger2007}, referred to as LASER in the following. It requires predicting the next value of a chaotic timeseries generated by the intensity pulsations of a real laser. The dataset consists of a sequence of $10092$ elements, normalized in $[-1,1]$, where the first $2000$ samples form the training sequence, the following $5000$ form the validation sequence and the remaining $3092$ are the test sequence. In both NARMA and LASER datasets, the first 200 samples of training, validation and test subsequences were used as initial washout period.

The experimental analysis focuses on assessing the effect of two topological features of the concentric ESN:
\begin{enumerate}
  \item the modular reservoir organization introduced by having multiple simple cycle groups of reservoir neurons;
  \item the effect of introducing jump connections between the simple cycle groups.
\end{enumerate}
To this end, we confront the empirical performance of the concentric ESN with jumps (cjESN) and without jumps (cESN), also with respect to their flat counterparts in literature, that are the SCR \cite{scr}  and the CRJ \cite{crj}. In order to better understand the impact of the modular organization, we confront the performance of the different models for the same number of total reservoir neurons $N_R$ and using the same hyperparameter values in the model selection procedure. Table \ref{tab:modsel} provides a summary of the hyperparameters considered in model selection.
\begin{table}[tb]
   \renewcommand{\arraystretch}{1.3}
   \caption{Network hyperparameters considered in model selection, including weights for the reservoir cycles $w_c$ and jumps $w_j$ (only for CRJ and cjESN), the rigde regression parameter $\lambda$, the leaking rate $\alpha$ and the jump size $\tau_j$.}\label{tab:modsel}
   \centering
  \begin{tabular}{|l|c|}
  \hline
  Parameter & Values\\
  \hline
  Input Weights &	$\{.1,.2,.3,.4,.5\}$\\
  \hline
  Cycle Weights $w_c$ &	$\{.4,.5,.6,.7,.8,.9,1\}$\\
  \hline
  Jump Weights $w_j$ &	$\{.1,.2,.3,.4,.5,.6,.7,.8,.9,1\}$\\
  \hline	
  $N_R$ &	$\{100,150,200,300,350,600\}$\\
  \hline
  $\lambda$ &	$\{10^k | k \in [-15,0]\}$\\
  \hline
  $\alpha$ &	$\{.1, .2, .5, .6, .7, .8, .9, 1\}$\\		
  \hline
  Jump Size	$\tau_j$ &	$\{5, 10, 15, 20, 30, 45\}$\\		
  \hline				
 \end{tabular}
\end{table}

In the modular cESN and cjESN models, one can have different reservoir organizations for the same total number of reservoir neurons, depending on the number of simple cycles as well as on their length. Table \ref{tab:topol} summarizes the different reservoir topologies considered in this analysis: note that the reservoir topology is itself an hyperparameter, hence we have chosen this on the validation performance as any other hyperparameter.   The choice of the possible reservoir configurations has been limited to a maximum of $3$ concentric reservoirs as well as to a fixed number of stencil configuration to avoid a combinatorial explosion in the configurations under test. Nevertheless, we believe the chosen topologies to be good representatives of the reservoir configurations, allowing to test: (i) simple cycles with equal size; (ii) simple cycles with increasing size from the outermost to the inner most; (iii) simple cycles with decreasing size from the outermost to the inner most; (iv) simple cycles with a bottleneck (i.e. less number of neurons) in the intermediate circle.
\begin{table*}[tb]
   \renewcommand{\arraystretch}{1.3}
   \caption{Reservoir topologies considered in model selection for the cESN and cjESN networks. Notation $n_1$-$n_2$-$n_3$ denotes a reservoir with a total of $N_R = n_1+n_2+n_3$ neurons, organized in 3 simple cycles where the outermost has $n_1$ neurons, the intermediate comprises $n_2$ neurons and the innermost has length $n_3$. Reservoir are constructed deterministically: hence, no multiple initialization is required to account for randomization.}\label{tab:topol}
   \centering
  \begin{tabular}{|l|c|}
  \hline
  $N_R$ & Topologies\\
  \hline
  100 &	50-50, 60-40, 40-60\\
  150 &	50-50-50, 75-75, 90-60\\
  200 &	100-50-50, 50-100-50, 50-50-100, 100-100\\						
  300 &	100-100-100, 150-100-50, 150-50-100, 50-150-100, 50-100-150, 100-50-150, 100-150-50\\						350 &	100-200-50, 100-50-200, 50-100-200, 50-200-100, 200-100-50, 200-50-100\\
  600 &	200-200-200, 300-300, 400-200, 200-400, 150-150-150-150\\		
  \hline				
 \end{tabular}
\end{table*}

The performance results of the models under test is show in Table \ref{tab:narma} for the NARMA dataset  and in Table \ref{tab:laser} for the LASER task. The first thing to note is that the jump connections have a positive effect on the modular architectures, confirming what has already been shown in \cite{crj} for the flat models. The cjESN model achieves test errors that are always smaller than those achieved by the equivalent cESN architecture. In the LASER task, in particular, the errors of the cjESN are constantly halved with respect to those by cESN. \begin{table}[tb]
   \renewcommand{\arraystretch}{1.3}
   \caption{Validation and test errors on NARMA, for varying reservoir sizes $N_R$,  for the best configuration of hyperparameters and topology selected in validation (lowest test errors are highlighted in bold).}\label{tab:narma}
   \centering
  \begin{tabular}{|l|l|c|c|c|c|}
  \hline
  $N_R$ & & SCR & cESN & CRJ & cjESN\\
  \hline
  \multirow{ 2}{*}{100} & Valid & 0.0007 & 0.0928	& 0.0119 & 0.1104\\
  & Test & 0.1137 & 0.1089	& 0.1086 & \bf{0.0797}\\
  \hline
  \multirow{ 2}{*}{150} & Valid & 0.0997 &	0.1056 & 0.0722 & 0.0661\\
  & Test	 & 0.0967 & 	0.0775 & 	\bf{0.0437} & 	0.0649\\
  \hline
  \multirow{ 2}{*}{200} & Valid & 0.0794 &	0.0742 & 0.0610	& 0.0449\\
  & Test & 	0.0544 & 	0.0481 & 	0.0356 & \bf{0.0334}\\
  \hline
  \multirow{ 2}{*}{300} & Valid & 0.0597 &	0.0589 & 0.0284	& 0.0261\\
  & Test & 	0.0313 & 0.0306 & 0.0303 & \bf{0.0188}\\
  \hline
  \multirow{ 2}{*}{350} & Valid & 0.0553 &	0.0560 & 0.0350 &	0.0289\\
  & Test & 	0.0257 & 0.0287 & 0.0175 & \bf{0.0164}\\
  \hline
  \multirow{ 2}{*}{600} & Valid & 0.0418 & 0.04168 & 0.01943 &	0.019\\
  & Test & 	0.0173	 & 0.0174 & 0.0143 & \bf{0.0137}\\
  \hline
 \end{tabular}
\end{table}

When confronting the performances between the modular architectures and the flat ones, one can note that the cjESN is almost always the best performing model (except for a single configuration in the NARMA task). By taking a closer look at the performances of cjESN and CRJ, it is clear how the modular topology of cjESN is particularly effective for smaller reservoirs (roughly up to $200$-$300$ neurons) where the performance differences between the two are particularly wide. With larger reservoirs (i.e. 600 neurons) such difference tends to fade, although still achieves the lowest errors. On the other hand, when comparing the two {\it jump-less} architectures, i.e. SCR and cESN, there seems to be a less clear distinction between the performance of the two models. For some configurations the modular architecture slightly prevails, while in others the SCR model has better performances, hence suggesting that the concentric topology needs jump connections in order to fully exploit its potential.
\begin{table}[tb]
   \renewcommand{\arraystretch}{1.3}
   \caption{Validation and test errors on LASER, for varying reservoir sizes $N_R$,  for the best configuration of hyperparameters and topology selected in validation (lowest test errors are highlighted in bold).}\label{tab:laser}
   \centering
  \begin{tabular}{|l|l|c|c|c|c|}
  \hline
  $N_R$ & & SCR & cESN & CRJ & cjESN\\
  \hline
  \multirow{ 2}{*}{100} & Valid & 0.0115	& 0.0122 & 0.0079	& 0.0117\\
  & Test & 0.0227 & 0.0215	& 0.0108 & \bf{0.0100}\\
  \hline
   \multirow{ 2}{*}{150} & Valid & 0.0105	& 0.0089 & 0.0058	& 0.0058\\
  & Test & 0.0160 & 0.0228	& 0.009 & \bf{0.0068}\\
  \hline
   \multirow{ 2}{*}{200} & Valid & 0.0101	& 0.0096 & 0.0054	& 0.0057\\
  & Test & 0.0119 & 0.0105 & 	0.0103 & \bf{0.0060}\\
  \hline
   \multirow{ 2}{*}{300} & Valid & 0.0084	& 0.0194 & 0.0048	& 0.0037\\
  & Test & 0.0119 & 0.0105 & 0.0060 & \bf{0.0047}\\
  \hline
   \multirow{ 2}{*}{350} & Valid & 0.0087	& 0.0084 & 0.0048	& 0.0045\\
  & Test & 0.0106 & 0.01 & 0.0077 & \bf{0.0049}\\
  \hline
   \multirow{ 2}{*}{600} & Valid & 0.0092	& 0.0087 & 0.0034	& 0.0049\\
  & Test & 0.0107	 & 0.0108 & 0.0061 & \bf{0.0052}\\
  \hline
 \end{tabular}
\end{table}

In order to better understand the effect of the topology, we report in Table \ref{tab:topo_narma} and \ref{tab:topo_laser} the reservoir configurations and the jump sizes selected for each reservoir size by the model selection process.  First, one can note how the cjESN tends to select pyramidal reservoirs (i.e. with decreasing cycle size) when the total number of neurons is smaller. Whereas, when the reservoir sizes increases, the preferred topologies tend to include either bottleneck architectures or fewer cycles with equal length. Since the configurations on which cjESN shows a clearer performance increase with respect to CRJ are those for smaller reservoir sizes, one might argue that topologies with decreasing number of neurons (as we move from outer-most to inner-most cycles) are better performing that other architectural types. In terms of jump size, the results in both tables show how cjESN tends to select more densely interconnected concentric cycles, while CRJ prefers  sparser jumps within its single cycle reservoir.
\begin{table}[tb]
   \renewcommand{\arraystretch}{1.3}
   \caption{Topological factors for architectures selected in NARMA validation}\label{tab:topo_narma}
   \centering
  \begin{tabular}{|cc|c|cc|}
  \hline
  \multicolumn{2}{|c|}{CRJ}  &  cESN & \multicolumn{2}{|c|}{cjESN}\\
  $N_R$  & $\tau_j$ & Topology & Topology & $\tau_j$\\
  \hline
  100 & 5  & 40-60 & 60-40 & 5\\
  150 & 15 & 50-50-50 & 90-60 & 5\\
  200 & 15 & 100-50-50 & 100-50-50 & 5\\
  300 & 30 & 150-100-50 & 150-50-100 & 15\\
  350 & 15 & 100-200-50 & 100-200-50 & 5\\
  600 & 30 & 200-400 & 300-300 & 15\\
  \hline
 \end{tabular}
\end{table}
\begin{table}[tb]
   \renewcommand{\arraystretch}{1.3}
   \caption{Topological factors for architectures selected in LASER validation}\label{tab:topo_laser}
   \centering
  \begin{tabular}{|cc|c|cc|}
  \hline
  \multicolumn{2}{|c|}{CRJ}  &  cESN & \multicolumn{2}{|c|}{cjESN}\\
  $N_R$  & $\tau_j$ & Topology & Topology & $\tau_j$\\
  \hline
  100 & 30  & 40-60 & 60-40 & 5\\
  150 & 15 & 90-60 & 90-60 & 5\\
  200 & 15 & 100-50-50 & 100-50-50 & 5\\
  300 & 5 & 150-50-100 & 150-50-100 & 5\\
  350 & 5 & 200-50-100 & 100-50-200 & 5\\
  600 & 30 & 200-400 & 300-300 & 15\\
  \hline
 \end{tabular}
\end{table}

\subsection{Memory Capacity} \label{sect:mc}

Memory Capacity (MC) has been introduced to characterize the (short term) memorization capabilities of a reservoir by measuring the ability of a network in encoding past events in their state space so that past values of an
i.i.d. input sequence can be recalled \cite{jaeger2002}. Estimation of MC is based on taking a univariate input sequence $u(t)$ as input to the network and, for a given delay $k$, train the network to output $u(t-k)$ after having seen the input stream $u(t-k) \dots u(t-1)u(t)$ up to time $t$.  Memory capacity at delay $k$ is computed as the squared correlation coefficient between the desired output $u(t-k)$ and the observed network output $y(t)$, i.e.
\begin{equation}
\label{eq.mck}
\mathbf{MC}_{k} = \frac{Cov^2\left(u(t-k),y(t)\right)}{Var(u(t))Var(y(t))}
\end{equation}
where $Cov$ and $Var$ are the covariance and variance operators, respectively. The short term MC is then obtained as
\begin{equation}
\label{eq.mc}
\mathbf{MC} = \sum_{k=1}^{\infty} MC_k.
\end{equation}
The MC of an ESN is typically estimated empirically by generating a sufficiently long stream of i.i.d data, training a readout with a neuron for each different delay $k$, up to a finite maximum delay $K$. In this paper, we use the same setting and the same data\footnote{Available here: \url{https://sites.google.com/site/cgallicch/resources}} used by the authors of \cite{deepESN}, so to be able to have a direct means of confrontation with a standard ESN as well as with more complex layered reservoirs such as in the Deep ESN model \cite{deepESN}.

The input stream contains 6000 observations, such that the first 5000 samples have been used for training and the remaining 1000 for test. We have considered delays up to $K=200$ and a total number of reservoir neurons $N_R=100$, which is also theoretical bound on the MC \cite{jaeger2002}. Following \cite{deepESN}, the input scaling has been fixed to $0.1$, the leaky parameter is chosen in $\alpha \in \{0,0.55,1\}$ and both the reservoir cycle weight and jump connection weight are taken from $\{0.1, 0.5, 0.9\}$. For the cjESN, we have considered jump steps $\tau \in \{5,25,50\}$ to test both very sparse and more dense cycle jumps.

We have tested both cESN and cjESN with two concentric cycles under different topologies to assess the effect on MC. Table \ref{tab:MCcjESN} summarizes the MC analysis for both architectures.  The jump-less cESN model seem more effective when the concentric reservoirs are organized into a pyramidal shape, i.e. where the number of neurons decreases when going from the outermost to the innermost cycle. The cjESN has again the best performance also in terms of memory capacity, but a pyramidal organization of the reservoir does not seem to be a key feature ensuring top-performances. Rather, the density of the jumps seems to be more critical, with the majority of the best performing models being characterized by more frequent jumps between cycles.
\begin{table}[tb]
   \renewcommand{\arraystretch}{1.3}
   \caption{Memory capacity of the cESN and cjESN for different topologies of the concentric reservoir (jump step for cjESN is between brackets).}\label{tab:MCcjESN}
   \centering
  \begin{tabular}{|l|c|c|}
  \hline
 Topology & cESN & cjESN\\
 \hline
 75-25 (5)	&   39.73	& 41.54\\
 25-75 (5)	& 	39.93	& 43.28\\
 40-60 (25)	& 	39.98	& 42.68\\
 20-80 (5)	& 	40.43	& \bf{43.75}\\
 15-85 (5) & 	41.01	& 43.55\\
 80-20 (5)	& 	41.04	& 43.62\\
 95-5 (25)	& 	41.31 & 41.52\\
 90-10 (5) & 	41.43 & 42.61\\
 85-15 (5) & 	41.80 & 42.82\\
  \hline
 \end{tabular}
\end{table}

Table \ref{tab:MCother} reports the MC of standard ESNs and of different deep ESN architectures on the same task, as described in \cite{deepESN}. One can note how two-cycles concentric networks yield considerably higher MC than the standard ESN and of the majority of deep architectures. In particular, only the deepESN model \cite{deepESN}, with $10$ layers of $10$ neurons and fixed $\alpha$ and spectral radius among layers, has levels of MC that are comparable to those of the concentric ESN. In particular, the best performing cjESN configuration shows an higher memory capacity than deepESN. One interesting aspect to note is the fact that concentric networks have a notably higher memory capacity that the grouped ESN and deepESN-IA model. The former is, in fact, very akin to a cESN, with its independent groups of reservoir neurons \cite{deepESN}. The latter, instead, is quite similar to a cjESN, with its stacked reservoirs, each receiving input from the previous reservoir layer plus data from the input neurons. The apparent simplicity of the concentric cycles allows to efficiently memorize information at different timescales using the dynamic memory of loops of different length, yielding to excellent results in terms of memory capacity.
\begin{table}[tb]
   \renewcommand{\arraystretch}{1.3}
   \caption{Memory capacity of a standard ESN and of different deep ESN architectures as described in \cite{deepESN}. DeepESN is characterized by a reservoir organized into successive layers, each receiving input only from the previous reservoir in the stack. The variant marked as + var $\alpha$ and $\rho$ allow the leaking parameter and spectral radius to vary in the different reservoir layers. The deepESN-IA adds the contribution of the input neurons to each reservoir layer, while a grouped ESN is a standard ESN with the reservoir partitioned into independent groups of neurons, each connected to the input neurons (but to the other reservoir modules).}\label{tab:MCother}
   \centering
  \begin{tabular}{|l|c|}
  \hline
  Model & MC\\
  \hline
standard ESN & 27.50\\
deepESN & \bf{42.45}\\
deepESN + var $\alpha$ & 37.15\\
deepESN + var $\rho$ & 30.79\\
deepESN-IA & 28.05\\
grouped ESN & 28.02\\
  \hline
\end{tabular}
\end{table}

\section{Conclusions} \label{sect:conclude}
We have proposed a novel modular architecture for ESN reservoirs that is based on a concentric organization of the reservoir neurons in unidirectional cycles linked by jump connections between the loops. The proposed approach relaxes the constrained SCR and CRJ architectures by introducing the possibility of explicitly modeling dynamic memory paths of different length as unidirectional cycles comprising a different number of units. The bidirectional jump connections, in this context, allow switching from one memory regime to the other by moving up and down in the concentric reservoir.

The experimental analysis has highlighted how the concentric architecture leads to increased predictive accuracy with respect to an equivalent reservoir organized following the SCR and CRJ layout. In particular, the results suggest that the concentric organization is effective even for smaller sizes of the reservoir, especially if jump connections are included.  The proposed approach has also been confronted with standard ESNs and with state of the art deepESN \cite{deepESN} in terms of empirical memory capacity, showing a significant advantage of the concentric reservoirs even when compared with the more articulated deepESN architectures.

The analysis performed in this paper is only preliminary and there are several directions for future works. In our simplified setting we have considered only concentric reservoirs having the same weight on each cycle. Allowing a selection of different weights for each cycle will permit to control the tradeoff between dynamic memory length (regulated by the number of units) and the speed of forgetting of recent history (regulated by the spectral properties of the cycle). This aspect would, in principle, allow to naturally process timeseries with diverse timescales within the same sequence, permitting a full exploitation of the jump connectivity between cycles. Another very interesting direction of future research concerns the introduction of a certain extent of adaptivity in the reservoir weights. In particular, it would be interesting to learn the values of the jump weights along the lines of what it has been done in \cite{crjAdapt} for the CRJ architecture. Another option to explore would be that of applying intrinsic plasticity mechanisms to the different reservoir cycles, as done in the deepESN model with excellent results in terms of memory capacity enhancement \cite{deepESN}.

\section*{Acknowledgment}
D. Bacciu is supported by the Italian Ministry of Education, University, and Research (MIUR) under project SIR 2014 LIST-IT (grant n. RBSI14STDE).

%\bibliographystyle{IEEEtran}
%\bibliography{circularESN}


\end{document}